\documentclass[conference]{IEEEtran}
\IEEEoverridecommandlockouts

\usepackage{cite}
\usepackage{amsmath,amssymb,amsfonts}
\usepackage{algorithmic}
\usepackage{graphicx}
\usepackage{textcomp}
\usepackage{xcolor}

\usepackage{amsmath}
\usepackage{booktabs}
\usepackage{graphicx}
\usepackage{url}

\usepackage[T1]{fontenc}   
\usepackage{listings}
\usepackage{mdframed}
\usepackage{enumitem}
\usepackage{multirow} 

\usepackage{tcolorbox}
\usepackage{pifont}

\newcommand{\cmark}{\ding{51}}
\newcommand{\xmark}{\ding{55}}

\def\BibTeX{{\rm B\kern-.05em{\sc i\kern-.025em b}\kern-.08em
    T\kern-.1667em\lower.7ex\hbox{E}\kern-.125emX}}

\newif\ifarxivversion
\arxivversionfalse

\begin{document}

\title{RDL-Bench: A Benchmark for Task-Aware Question Answering over Relational Databases
}


\author{\IEEEauthorblockN{Yao Cheng}
\IEEEauthorblockA{\textit{College of Computing and Data Science} \\
\textit{Nanyang Technological University}\\
Singapore \\
yao.cheng@ntu.edu.sg}
\and
\IEEEauthorblockN{Siqiang Luo}
\IEEEauthorblockA{\textit{College of Computing and Data Science} \\
\textit{Nanyang Technological University}\\
Singapore \\
siqiang.luo@ntu.edu.sg}
}
\maketitle

\begin{abstract}
Relational databases (RDBs) remain the primary 
infrastructure for storing and managing structured data 
across enterprise and scientific domains.
Natural-language (NL) interfaces to RDBs aim to let 
users ask questions without writing SQL, but real user 
needs go beyond SQL-style lookup: some questions require 
reasoning over the observable database state, while 
others require predictive inference from historical 
relational records under temporal constraints.
Existing benchmarks address these goals in isolation.
To address this gap, we introduce RDL-Bench, a benchmark for task-aware question answering (QA) over real RDBs.
RDL-Bench organizes diverse user questions into reasoning-oriented questions, whose answers are determined by the current database state, and prediction-oriented questions, whose answers depend on future outcomes under temporal constraints.
Each instance contains an NL question, task metadata, schema information, temporal context, either executable evidence or target-construction logic, and denotation-level ground truth.
RDL-Bench supports two evaluation settings: native solvers under their intended capabilities, and task-aware modular systems where a router dispatches each question to an appropriate execution module.
Experiments across seven domains yield several key findings.
Current reasoning methods achieve at most 0.49 
macro-average denotation score, with context augmentation 
providing no consistent improvement, revealing that schema 
grounding, relational composition, and answer normalization 
remain jointly unsolved.
LLM predictive competitiveness is format-driven and 
collapses on scalar regression, showing that binary 
predictive QA overestimates LLM relational understanding.
Task-aware routing exposes a capability threshold: large 
LLMs route reliably while small LLMs develop systematic 
bias that degrades end-to-end performance below random 
baseline.
Datasets and code are available at
\url{https://github.com/cy623/RDL-Bench}.
\end{abstract}


\section{Introduction}
\label{sec:intro}

Relational databases (RDBs) remain the core infrastructure for storing and managing structured data in domains such as finance, healthcare, e-commerce, transportation, social platforms, and public services~\cite{codd1970relational,stonebraker2010sql}.
As natural-language (NL) interfaces to RDBs mature, users increasingly expect to ask diverse questions over the same database without writing Structured Query Language (SQL).
These questions go beyond SQL-style lookup.
Some ask what is true in the current database state, such as factual retrieval, analytical aggregation, or multi-hop relational reasoning.
Others ask what is likely to happen in the future, such as future counts, labels, scores, or rankings inferred from historical relational records.
For example, a user may ask which cities had the most customers in a given period, or whether a seller is likely to receive low ratings next month.
Both are NL questions over relational data, but answering them requires different forms of database grounding, temporal reasoning, and task execution.

Current-state query answering and future-oriented prediction
are studied in text-to-SQL~\cite{yu2018spider,li2023can}
and relational-learning benchmarks~\cite{3}, respectively.
This motivates the problem of \textit{task-aware relational 
QA}, where a system must not only produce an answer but 
also identify the appropriate execution mechanism from the question alone.
In realistic user-facing settings, users do not pre-declare 
whether a question should be answered by querying the 
observable database state or by making a future-oriented 
prediction.
Moreover, the same user may ask both types of questions in 
sequence over the same database.
A practical relational QA system must therefore understand 
the user intent, ground the question in the schema and 
temporal context, identify the capability required to answer 
it, and invoke an appropriate execution mechanism.
The key challenge is thus not only producing an answer, but 
also \textit{deciding how the question should be answered}.

Existing benchmarks cover important parts of this broader goal, but mostly in isolation.
Table question answering (Table QA) and text-to-SQL benchmarks evaluate NL access to structured data, typically by answering questions over existing database states or generating executable SQL queries~\cite{pasupat2015compositional,zhongSeq2SQL2017,yu2018spider,li2023can}.
Recent complex analytical QA benchmarks further stress reasoning over structured databases, including nested queries, time-series analysis, and computations over aggregate results~\cite{maddela2025starqa}.
In parallel, relational deep learning (RDL) benchmarks such as RelBench~\cite{3} evaluate supervised prediction over relational databases.
However, these benchmark families do not provide a unified user-facing workload in which NL questions over the same type of relational data may require either current-state reasoning or future-oriented prediction.
In these settings, the required capability is typically specified in advance, leaving the system-level impact of inferring task requirements outside the evaluation.

\begin{figure}[t]
\centering
\includegraphics[width=1\linewidth]{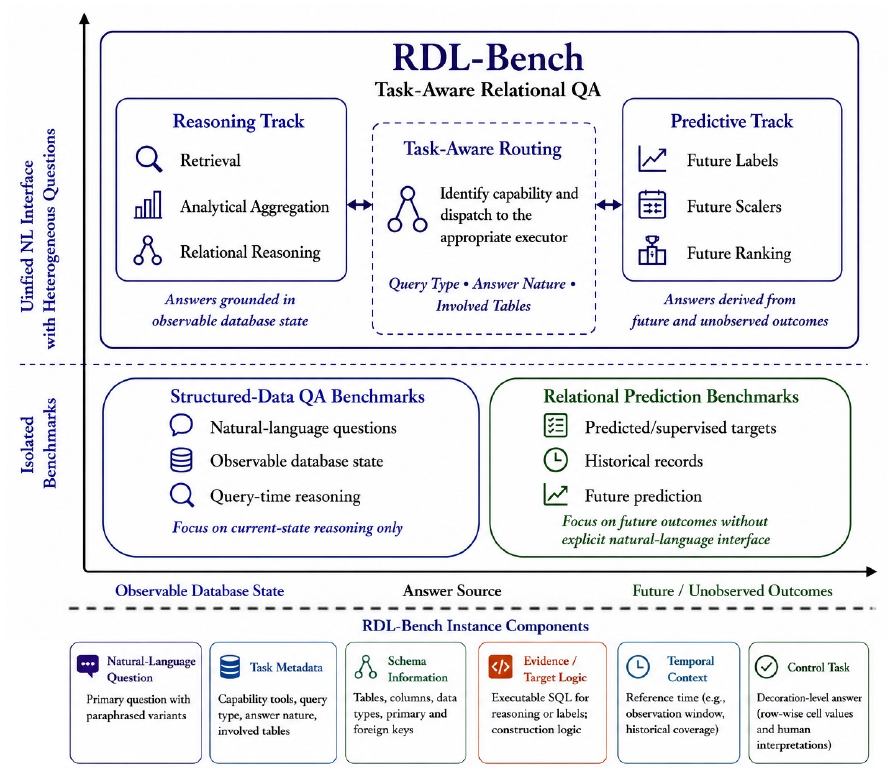}
\vspace{-2em}
\caption{Overview of RDL-Bench in the relational QA landscape.
RDL-Bench places current-state reasoning and future-oriented prediction under a shared NL interface and evaluates both native solvers and task-aware modular systems.
Each instance contains six components: NL question, task metadata, schema information, evidence or target-construction logic, temporal context, and denotation-level ground truth.}
\vspace{-1.5em}
\label{fig:rdlbench-overview}
\end{figure}

To address the above gap, we introduce \textbf{RDL-Bench}, a benchmark for task-aware QA over real multi-table RDBs.
Figure~\ref{fig:rdlbench-overview} summarizes the benchmark space considered in this work.
RDL-Bench is built around a unified NL interface: each instance is formulated as a user question over an RDB and evaluated against ground truth.
To support task-aware evaluation, each instance further includes task metadata, schema information, temporal context, and either executable evidence or target-construction logic.

RDL-Bench covers the core capabilities required by practical relational QA systems.
For reasoning-oriented questions, answers are determined by the observable database state, and ground truth is obtained by executing validated SQL queries.
For prediction-oriented questions, answers depend on outcomes after a reference time $t_{\mathrm{ref}}$; the input context is restricted to records observed before $t_{\mathrm{ref}}$, and labels are constructed under strict temporal constraints to avoid information leakage.
This design enables RDL-Bench to evaluate whether a system can understand the user question, identify the required capability, invoke the appropriate execution mechanism, and produce the correct answer.

RDL-Bench supports two evaluation settings.
The first setting measures \textit{native solvers} under their intended capabilities, including large language model (LLM)-based and text-to-SQL methods for reasoning-oriented questions, as well as tabular and graph-based RDL models for prediction-oriented questions.
The second setting assesses \textit{task-aware modular QA systems}, where a router predicts the required task route and dispatches the question to either a reasoning module or a prediction module.
By comparing an LLM-based router with an oracle router that uses ground-truth task metadata, RDL-Bench separates execution errors from routing errors and quantifies the system-level gap caused by imperfect task understanding.
Routing thus serves as a system-level diagnostic within the unified QA workload.
In summary, our contributions are as follows:

\begin{itemize}
    \item We introduce RDL-Bench, a benchmark for task-aware 
    QA over real multi-table RDBs under a unified NL 
    interface.

    \item We define an NL workload covering both 
    current-state reasoning and future-oriented prediction, 
    organized under a shared instance schema with task 
    metadata, temporal context, and denotation-level ground 
    truth.

    \item We introduce a two-level evaluation protocol 
    measuring native solver performance and modular QA 
    performance, including the routing-induced gap between 
    an LLM-based router and an oracle router, across seven 
    domains.
\end{itemize}

\noindent\textbf{Key Findings.}
Experiments across seven real-world domains yield
the following findings:
\begin{itemize}
\item \textbf{Relational reasoning remains limited by denotation alignment.}
The best seven-domain native reasoning macro-average
is 0.49, with performance varying across domains and
answer forms. Current methods struggle to align user
intent, schema grounding, relational composition,
and answer normalization.

\item \textbf{More relational context is not always sufficient.}
Two-hop context helps in some domains but does not
consistently improve over schema-only prompting,
indicating that more local records alone do not
resolve query planning and answer-form errors.

\item \textbf{LLM predictive performance is format-driven.}
LLMs can be competitive on Boolean prediction from
schema and historical statistics but remain unstable
on scalar regression, showing that Boolean predictive
QA can overestimate relational understanding.

\item \textbf{Routing separates schema grounding from capability understanding.}
Strong LLMs route accurately, whereas Llama8B can
identify relevant tables yet fail to distinguish
reasoning from prediction. Routing requires recognizing
the execution capability implied by the question.
\end{itemize}

\section{Related Work}
\label{sec:related_work}

\subsection{Question Answering over Structured Data}

Question answering over structured data has been studied 
through Table QA, semantic parsing, and NL interfaces to 
databases.
WikiTableQuestions~\cite{pasupat2015compositional} 
introduced complex QA over semi-structured Wikipedia tables,
requiring compositional operations such as comparison, 
aggregation, and arithmetic 
reasoning~\cite{chen2019tabfact}.
WikiSQL~\cite{zhongSeq2SQL2017,xu2017sqlnet} provided a 
large-scale dataset of NL questions paired with SQL queries 
over single tables.
Spider~\cite{yu2018spider} advanced the field with a 
complex cross-domain text-to-SQL benchmark over multi-table 
databases.
BIRD~\cite{li2023can} further emphasizes large-scale 
database-grounded text-to-SQL with realistic database 
values, external knowledge, and SQL efficiency.
Conversational database QA has been studied in 
CoSQL~\cite{yu2019cosql}, which focuses on multi-turn 
interactions grounded in SQL queries.
STARQA~\cite{maddela2025starqa} is the most closely related 
work, studying complex analytical reasoning over structured 
databases via Text2SQLCode, which combines SQL-based 
retrieval with Python-based post-processing.
However, all these benchmarks restrict questions to those 
whose answers are determined by the current database state, 
and none evaluates whether a system can recognize the 
required capability and route accordingly.
Table-Augmented Generation (TAG)~\cite{biswal2024text2sql} combines database execution with language-model reasoning to answer questions requiring semantic interpretation or world knowledge beyond conventional text-to-SQL.
RDL-Bench complements this direction by evaluating current-state reasoning and future-oriented prediction under a shared NL interface, with explicit temporal constraints for predictive tasks and separate evaluation of native solvers and task-aware QA systems.

\subsection{Text-to-SQL Methods}

Text-to-SQL methods translate NL questions into executable 
SQL queries~\cite{zhongSeq2SQL2017,guo2019towards}.
RAT-SQL~\cite{wang2020rat} introduced relation-aware schema 
encoding, and RESDSQL~\cite{li2023resdsql} decoupled schema 
linking from SQL skeleton parsing to improve robustness on 
complex queries.
With the emergence of LLMs, prompting-based methods such as 
DIN-SQL~\cite{pourreza2023din} and C3~\cite{dong2023c3} 
have shown strong performance by decomposing SQL generation 
or improving zero-shot prompting.
These methods are effective when the answer can be derived 
by executing SQL over an existing database snapshot, but are 
not designed for future-oriented prediction.
RDL-Bench evaluates these methods as native solvers for 
reasoning-oriented questions and characterizes their 
cross-domain limitations.

\subsection{Relational Deep Learning and Predictive 
Modeling}

Different from graph neural networks~\cite{song2021dynamic,song2024efficient,song2026ehrag,song2026mitigating,song2026signed,LiAhZh24,LiSuZe26,LiXiZe26,LiZhAh25,LiZhZh26}, RDL studies predictive modeling over relational data with 
multiple entity types, relations, and temporal 
dependencies~\cite{rossi2020temporal}.
A common approach transforms relational data into 
entity-level features for tabular learning methods such as 
XGBoost~\cite{chen2016xgboost} and 
LightGBM~\cite{ke2017lightgbm}.
Graph-based methods model relational structure directly: 
GraphSAGE~\cite{hamilton2017inductive} learns inductive node 
representations, while heterogeneous graph models such as 
HGT~\cite{hu2020heterogeneous} handle multiple node and 
edge types.
RelBench~\cite{3} provides standardized evaluation for 
supervised prediction over relational databases, focusing 
on Predictive Track defined over relational structure and 
temporal information.
RDL-Bench extends this line of work by expressing 
Predictive Track as NL user questions and evaluating them 
jointly with reasoning-oriented questions under a unified 
QA framework.

\begin{table}[t]
\centering
\caption{Comparison of related benchmarks. 
}
\vspace{-0.8em}
\label{tab:related}
\small
\resizebox{\columnwidth}{!}{%
\begin{tabular}{lcccc}
\toprule
Benchmark & Reasoning & Predictive & Routing 
& Multi-table \\
\midrule
WikiTableQ~\cite{pasupat2015compositional} 
& \cmark & \xmark & \xmark & \xmark \\
Spider~\cite{yu2018spider}     
& \cmark & \xmark & \xmark & \cmark \\
BIRD~\cite{li2023can}          
& \cmark & \xmark & \xmark & \cmark \\
STARQA~\cite{maddela2025starqa}
& \cmark & \xmark & \xmark & \cmark \\
RelBench~\cite{3}              
& \xmark & \cmark & \xmark & \cmark \\
\midrule
\textbf{RDL-Bench}             
& \cmark & \cmark & \cmark & \cmark \\
\bottomrule
\end{tabular}
}
\vspace{-2em}
\end{table}

\subsection{Positioning RDL-Bench}

Table~\ref{tab:related} summarizes representative
benchmarks and their coverage across key dimensions.
Structured-data QA benchmarks provide NL interfaces and
evaluate current-state reasoning, but do not support
Predictive Track or routing evaluation.
RDL benchmarks such as RelBench target future-oriented
prediction over relational structure, but without a
user-facing NL interface or routing component.
RDL-Bench brings current-state query answering and future-oriented prediction into a shared NL workload with a common instance schema.
Its two-level evaluation distinguishes native solver quality from end-to-end task-aware QA performance, with routing serving as a system-level diagnostic of errors in inferring the required capability.

\section{Preliminaries and Problem Definition}
\label{sec:problem_formulation}

We study QA over RDBs.
Given an NL question over an RDB, the goal is to produce the correct answer by identifying the capability required by the question and invoking an appropriate execution mechanism.
RDL-Bench does not assume that a single model family must solve all question types.
Instead, it evaluates both native solvers under their intended capabilities and modular systems that route heterogeneous user questions to specialized execution modules.

\vspace{-0.6em}
\subsection{Relational Databases and QA Instances}
\label{subsec:preliminaries}

Let $D=\{T_1,T_2,\ldots,T_K\}$ denote an RDB with $K$ tables, and let $S$ denote its schema.
The schema $S$ specifies table names, attributes, data types, primary keys, and foreign-key relations.
For a table $T_i \in D$, each tuple $\tau \in T_i$ contains a set of attribute values.
When available, a tuple may be associated with an event timestamp $t(\tau)$.
A benchmark instance is represented as:
\begin{equation}
\vspace{-0.5em}
    x=(D,S,q,t_{\mathrm{ref}},a),
\end{equation}
where $q$ is an NL user question, $t_{\mathrm{ref}}$ is an optional reference time, and $a$ denotes task metadata such as question type, answer form, target entity, prediction horizon, or involved schema components.
The target answer is denoted by $y$.
For non-temporal reasoning questions, $t_{\mathrm{ref}}$ may be absent.
For predictive questions, $t_{\mathrm{ref}}$ defines the boundary between observable context and future target outcomes.

We denote the observable database context by $D_{\mathrm{ctx}}$.
For reasoning-oriented questions, $D_{\mathrm{ctx}}$ is the database snapshot used to answer the question.
For prediction-oriented questions, $D_{\mathrm{ctx}}$ contains only tuples satisfying $t(\tau)\leq t_{\mathrm{ref}}$, while the target answer $y$ is constructed from records after $t_{\mathrm{ref}}$ according to the task metadata $a$.
This separation prevents temporal leakage in predictive evaluation.

\begin{table*}[h]
\centering
\caption{Statistics of the seven domains in RDL-Bench. Diversity is reflected in the varying number of tables, relational depth, and the presence of temporal event streams.}
\vspace{-1em}
\label{tab:db_stats}
\resizebox{\textwidth}{!}{
\begin{tabular}{@{}lcccccl@{}}
\toprule
\textbf{Name} & \textbf{Domain} & \textbf{\# Tables} & \textbf{\#Rows} & \textbf{Temporal} & \textbf{Source} \\
\midrule
\textbf{Olist} & E-commerce & 9 & 100,000 & Yes & \url{https://www.kaggle.com/datasets/olistbr/brazilian-ecommerce/data} \\
\textbf{Financial} & Financial & 8 & 1,090,086 & Yes & \url{https://web.archive.org/web/20180506061559/http://lisp.vse.cz/pkdd99/Challenge/chall.htm} \\
\textbf{NCAA} & Sports & 9 & 201,555 & Yes & \url{https://www.kaggle.com/c/march-machine-learning-mania-2015/data} \\
\textbf{Hepatitis} & Medicine & 7 & 12,927 & No & \url{https://www.cs.sfu.ca/~oschulte/jbn/dataset.html} \\
\textbf{Airline} & Retail & 19 & 448,156 & Yes & \url{https://www.transtats.bts.gov} \\
\textbf{Stats} & Social Network & 8 & 1,027,838 & Yes & \url{https://archive.org/details/stackexchange} \\
\textbf{Dallas} & Government & 3 &  812 & Yes & \url{https://www.dallasopendata.com/Public-Safety/Dallas-Police-Officer-Involved-Shootings/4gmt-jyx2/about_data} \\
\bottomrule
\end{tabular}
}
\vspace{-2.2em}
\end{table*}

\begin{figure}[h]
  \centering  \includegraphics[width=1\linewidth]{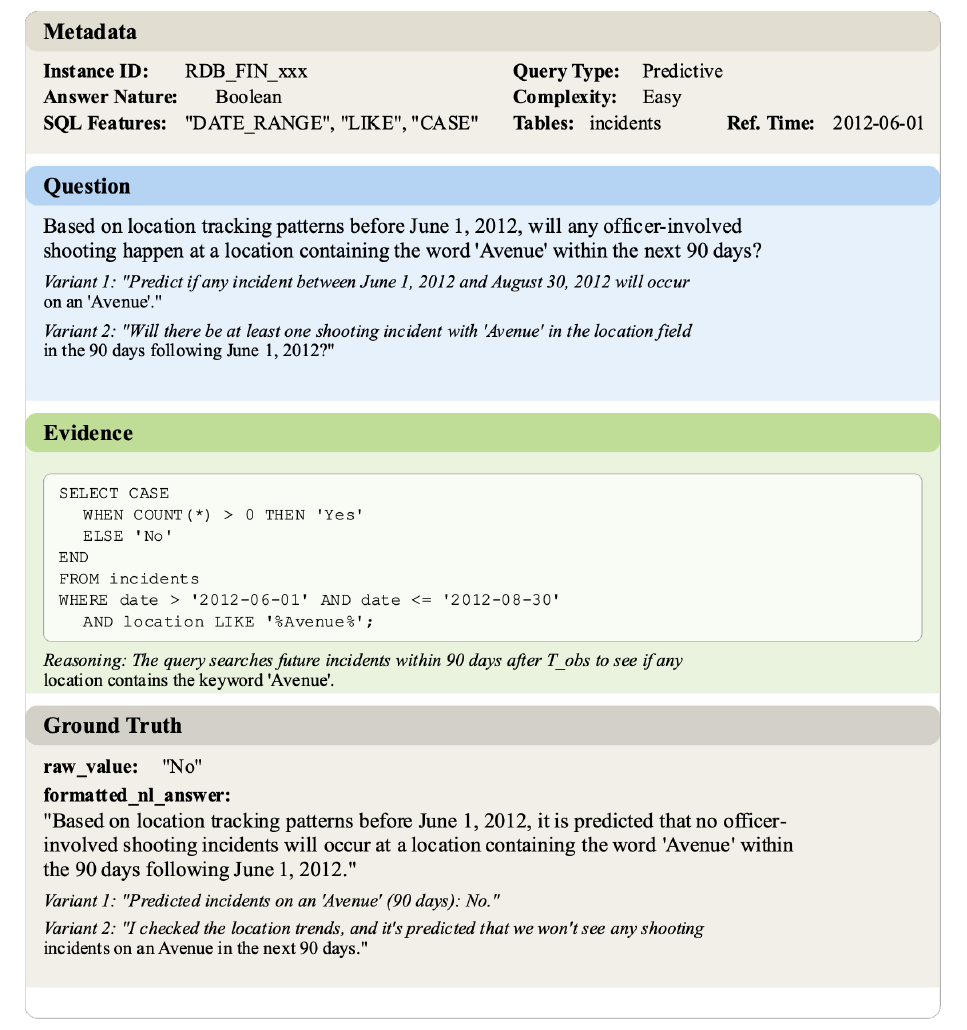}
  \vspace{-2.2em}
  \caption{Example of a predictive instance.
  }
  \vspace{-1.5em}
  \label{fg:task_instance}
\end{figure}

\subsection{Capability Tracks}
\label{subsec:capability_tracks}

RDL-Bench organizes questions into two capability tracks according to how their answers are determined.

\paragraph{Reasoning Track}
The Reasoning Track contains questions whose answers are determined by the observable database state.
It includes factual retrieval, analytical aggregation, and multi-hop relational reasoning.
Such questions can typically be answered through symbolic execution over $D_{\mathrm{ctx}}$:
\begin{equation}
\vspace{-0.5em}
    y = \operatorname{Exec}(D_{\mathrm{ctx}}, s_q),
\end{equation}
where $s_q$ denotes an executable query or program corresponding to the NL question $q$, and $\operatorname{Exec}$ denotes executing $s_q$ over $D_{\mathrm{ctx}}$ to obtain the answer.

\paragraph{Predictive Track}
The Predictive Track contains questions whose answers depend on unobserved outcomes.
Given the observable context $D_{\mathrm{ctx}}$ before $t_{\mathrm{ref}}$, the system must estimate a target constructed from records after $t_{\mathrm{ref}}$:
\begin{equation}
    y = h(D_{>t_{\mathrm{ref}}}, a),
    \vspace{-0.5em}
\end{equation}
where $h(\cdot)$ defines the task-specific target construction, such as a future label, score, scalar value, count, or ranking.
Unlike reasoning-oriented questions, predictive questions cannot be answered by directly querying $D_{\mathrm{ctx}}$; they require models that generalize from historical relational patterns.

\vspace{-0.5em}
\subsection{Task-Aware Relational QA}
\label{subsec:task_aware_system}

A task-aware relational QA system consists of a router and a set of execution modules.
Given a question $q$ and schema $S$, the router predicts a structured route:
\begin{equation}
    z = g(q,S),
    \vspace{-0.5em}
\end{equation}
where $z$ includes both the capability track and fine-grained routing metadata, such as the query type, answer form, target entity, and relevant schema components.

Routing involves both coarse-grained and fine-grained decisions.
At the coarse level, the router determines whether the question belongs to the Reasoning Track or the Predictive Track.
At the fine-grained level, it identifies the expected answer form and the schema components or target definition required by the execution module.
These decisions determine which module is invoked and how its output should be interpreted.
Routing errors may cause downstream failures even when the execution modules themselves are strong.

The system selects an execution module $m_z$ from a module set $\mathcal{M}$ and produces the final answer:
\begin{equation}
\vspace{0.5em}
    \hat{y} = m_z(D_{\mathrm{ctx}}, S, q, z).
    \vspace{-1em}
\end{equation}
In RDL-Bench, reasoning modules include LLM-based and text-to-SQL methods that generate executable queries, while predictive modules include tabular and graph-based RDL models trained on temporally valid historical data.
This formulation separates routing errors from execution errors.

\vspace{-0.5em}
\subsection{Evaluation Objective}
\label{subsec:objective}

RDL-Bench evaluates denotation-level correctness between the predicted answer $\hat{y}$ and the ground-truth answer $y$.
Because answer formats differ across question types, the loss function is type-aware:
\begin{equation}
    \mathcal{L}(\hat{y},y; a),
    \vspace{-0.5em}
\end{equation}
where the task metadata $a$ determines the evaluator used for the instance, such as exact match for scalar answers, execution-based correctness for executable queries, or task-specific metrics for predictive outputs.

The overall objective is to minimize the expected task loss over heterogeneous relational QA instances:
\begin{equation}
    \mathcal{J}
    =
    \mathbb{E}_{(x,y)\sim \mathcal{P}}
    \left[
        \mathcal{L}(\hat{y},y; a)
    \right],
    \vspace{-0.5em}
\end{equation}
where $\mathcal{P}$ denotes the benchmark distribution over databases, schemas, questions, capability tracks, task metadata, and answer types.

For native solvers, the route $z$ is fixed by the capability track being evaluated.
For task-aware modular systems, $z$ is predicted by the router.
This distinction allows RDL-Bench to measure both capability-specific execution performance and end-to-end task-aware QA performance.

\begin{figure}[h]
\centering
\begin{mdframed}[
    linewidth=0.6pt,
    linecolor=black!30,
    backgroundcolor=black!2,
    innerleftmargin=5pt,
    innerrightmargin=5pt,
    innertopmargin=4pt,
    innerbottommargin=4pt
]

\scriptsize
\setlength{\parskip}{0pt}
\setlength{\parindent}{0pt}

\centering
\textbf{Prompt Template A: Reasoning Track Instance Generation}
\par\vspace{2pt}

\raggedright
\textbf{[Role]}
You are an expert Data Scientist and Database Evaluation Specialist.
Generate temporally grounded questions for NL2SQL and RAG benchmark construction.

\vspace{2pt}

\begin{minipage}[t]{0.49\linewidth}
\textbf{[Query Categories]}
\begin{itemize}[leftmargin=1.1em,itemsep=0pt,topsep=1pt,parsep=0pt]
\item \textit{Retrieval}: data fetching via ranges, strings, and dates.
\item \textit{Analytical}: aggregation using \texttt{GROUP BY}, \texttt{HAVING}, and aggregate functions.
\item \textit{Relational Reasoning} [Hard]: multi-hop logic with \texttt{JOIN}s, subqueries, or set operations.
\end{itemize}
\end{minipage}
\hfill
\begin{minipage}[t]{0.49\linewidth}
\textbf{[Temporal Selection]}
\begin{itemize}[leftmargin=1.1em,itemsep=0pt,topsep=1pt,parsep=0pt]
\item \textit{Validity}: \texttt{time} must fall within table date ranges.
\item \textit{Realism}: avoid dates before relevant entities exist.
\item \textit{Diversity}: vary time points across instances.
\end{itemize}
\end{minipage}

\vspace{2pt}

\textbf{[Output Schema]}

\vspace{2pt}

\textbf{[Quality Constraints]}
\begin{itemize}[leftmargin=1.1em,itemsep=0pt,topsep=1pt,parsep=0pt]
\item Sequential IDs from \texttt{RDB\_\{DOMAIN\}\_\{START\}}, incremented by 1.
\item No repeated logical patterns or table combinations across batches.
\item Only use tables and columns defined in the provided schema.
\item No hallucinated foreign-key relationships.
\item \texttt{nl\_variants} must reflect real-world user phrasing.
\item Do not query values with 0\% distribution in the profiling data.
\end{itemize}

\end{mdframed}
\vspace{-1em}
\caption{Generation prompt template for Reasoning task instances.}
\vspace{-1em}
\label{fig:prompt-reasoning}
\end{figure}

\begin{figure}[t]
\centering
\begin{mdframed}[
    linewidth=0.6pt,
    linecolor=black!30,
    backgroundcolor=black!2,
    innerleftmargin=5pt,
    innerrightmargin=5pt,
    innertopmargin=4pt,
    innerbottommargin=4pt
]

\scriptsize
\setlength{\parskip}{0pt}
\setlength{\parindent}{0pt}

\centering
\textbf{Prompt Template B: Predictive Track Instance Generation}
\par\vspace{2pt}

\raggedright
\textbf{[Role]}
You are a Data Scientist. Generate \textit{prospective} questions that require
a model to analyze historical patterns before $T_{\mathrm{obs}}$ and predict
future events after $T_{\mathrm{obs}}$.

\vspace{2pt}

\textbf{[Temporal Boundary]}
$T_{\mathrm{obs}}$ defines a strict causal wall: data after
$T_{\mathrm{obs}}$ is invisible during inference, while \texttt{label\_sql}
queries only future records to construct the ground-truth label.

\vspace{2pt}

\begin{minipage}[t]{0.49\linewidth}
\textbf{[Temporal Parameters]}
\begin{itemize}[leftmargin=1.1em,itemsep=0pt,topsep=1pt,parsep=0pt]
\item \texttt{reference\_time}: the value of $T_{\mathrm{obs}}$.
\item \texttt{prediction\_window} $(\Delta t)$: future duration being predicted.
\item \texttt{signal\_period}: historical look-back window before $T_{\mathrm{obs}}$.
\end{itemize}
\end{minipage}
\hfill
\begin{minipage}[t]{0.49\linewidth}
\textbf{[Quality Constraints]}
\begin{itemize}[leftmargin=1.1em,itemsep=0pt,topsep=1pt,parsep=0pt]
\item \textit{Causal Wall}: \texttt{primary\_nl} asks only about the future.
\item \textit{Answer Type}: binary or quantitative answers.
\item \textit{Feasibility}: use only the provided schema.
\item \textit{No Peeking}: avoid future events in \texttt{primary\_nl}.
\end{itemize}
\end{minipage}

\vspace{2pt}

\textbf{[Output Schema]}

\end{mdframed}
\vspace{-1em}
\caption{Generation prompt template for Predictive task instances.}
\vspace{-2em}
\label{fig:prompt-predictive}
\end{figure}

\vspace{-0.5em}
\section{Dataset Construction}
\label{sec:dataset_construction}

RDL-Bench is constructed to support systematic evaluation 
of task-aware QA over real RDBs.
The benchmark follows three design principles that 
directly enable the experimental analyses in 
Section~\ref{sec:experiments}.
First, \emph{realism}: all questions are grounded in real 
multi-table RDBs with non-trivial schemas and foreign-key 
structures, ensuring that schema grounding and relational 
composition are genuinely required.
Second, \emph{consistency}: all instances share a unified 
JSON schema across domains and task types, enabling 
fair cross-domain and cross-type comparisons.
Third, \emph{verifiability}: each instance is associated with executable evidence or temporally grounded target-construction
logic, enabling denotation-level evaluation.

\vspace{-0.8em}
\subsection{Domains and Data Sources}
\label{subsec:domains}


RDL-Bench is built from seven real-world RDBs spanning
e-commerce, finance, sports, healthcare, transportation, social
platforms, and government. The selected databases vary in
schema size, number of tables, relational depth, temporal cov-
erage, and entity distributions. Each database contains multiple
interrelated tables with realistic schema designs, including
primary keys, foreign-key relations, temporal attributes, and
domain-specific entity types. This diversity is important for
evaluating relational QA under heterogeneous database work-
loads. A system must process unseen schemas, identify task-relevant tables and attributes, and handle different relational
structures across domains. 
Summary statistics of the databases
are reported in Table~\ref{tab:db_stats}.

\vspace{-0.7em}
\subsection{Task Taxonomy}
\label{subsec:task_taxonomy}

RDL-Bench organizes instances according to the capability required to answer the question.

\paragraph{Reasoning-oriented questions}
Reasoning-oriented questions are answered from the observable database state.
They include three task categories.

\begin{itemize}
    \item \textbf{Retrieval.}
    Retrieval tasks require selecting records or attributes that satisfy explicit conditions.
    They typically involve filtering, projection, and simple joins.

    \item \textbf{Analytical.}
    Analytical tasks require aggregation, comparison, or ranking over relational data.
    Common operations include \texttt{GROUP BY}, \texttt{COUNT}, \texttt{COUNT DISTINCT}, \texttt{AVG}, \texttt{ORDER BY}, and \texttt{LIMIT}.

    \item \textbf{Relational reasoning.}
    Relational Reasoning Track require relational composition across multiple tables or conditions.
    They may involve multiple joins, nested queries, set operations, or indirect constraints.
\end{itemize}

\paragraph{Prediction-oriented questions}
Prediction-oriented questions ask about future outcomes.
Each predictive instance is associated with a reference time $t_{\mathrm{ref}}$ and, when applicable, a prediction window $\Delta t$.
The observable context contains only records available before $t_{\mathrm{ref}}$, while the target answer is derived from records after $t_{\mathrm{ref}}$.
This setting evaluates whether a predictive model can estimate future labels, scores, scalar values, counts, or rankings from historical relational context.

\vspace{-1em}
\subsection{Unified Instance Schema}
\label{subsec:instance_schema}
Despite differences across domains and task types, all benchmark instances are represented using a shared JSON-based schema.
Each instance corresponds to an NL user question over an RDB and contains six logical components: the question, task metadata, schema information, evidence or target-construction logic, temporal context, and denotation-level ground truth.
Figure~\ref{fg:task_instance} shows an example.

\paragraph{Question}
The question component contains a primary NL question and multiple paraphrased variants.
These variants support evaluation under linguistic variation while preserving the same underlying relational task.

\paragraph{Task metadata}
The metadata component specifies the task type, answer form, complexity level, SQL features, involved tables, and reference time when applicable.
The \texttt{query\_type} field identifies whether the instance is retrieval, analytical, relational reasoning, or predictive.
The \texttt{answer\_nature} field specifies the expected answer form, such as scalar, set, boolean, tuple collection, or ranking.
The complexity fields record structural properties of the task, including required SQL operators and tables involved.

\paragraph{Schema information}
The schema component records the database schema relevant, including table names, attributes, data types, primary keys, and foreign-key relations.
This information supports schema grounding and allows systems to identify the database elements required by the question.

\paragraph{Evidence}
For reasoning-oriented questions, this component contains a validated SQL query and a short description of the corresponding relational operations.
For prediction-oriented questions, it specifies the target construction logic, reference time, prediction window, and label-generation rule used to derive the answer from future records.

\paragraph{Temporal context}
The temporal context specifies the time range relevant to the instance.
For prediction-oriented questions, it defines the observable context before $t_{\mathrm{ref}}$ and the future window used for target construction.
For reasoning-oriented questions, temporal information is included when the question itself contains temporal constraints.

\paragraph{Ground truth}
The ground-truth component stores the denotation-level answer in \texttt{raw\_value}.
Depending on the task, the raw value may be a scalar, set, boolean, tuple collection, or ranking.
For readability, each instance also includes a formatted NL answer and optional answer variants.
These formatted answers are used for human interpretation and auxiliary analysis, while primary evaluation is based on \texttt{raw\_value}.

\begin{figure}[t]
\vspace{-0.5em}
  \centering  \includegraphics[width=1\linewidth]{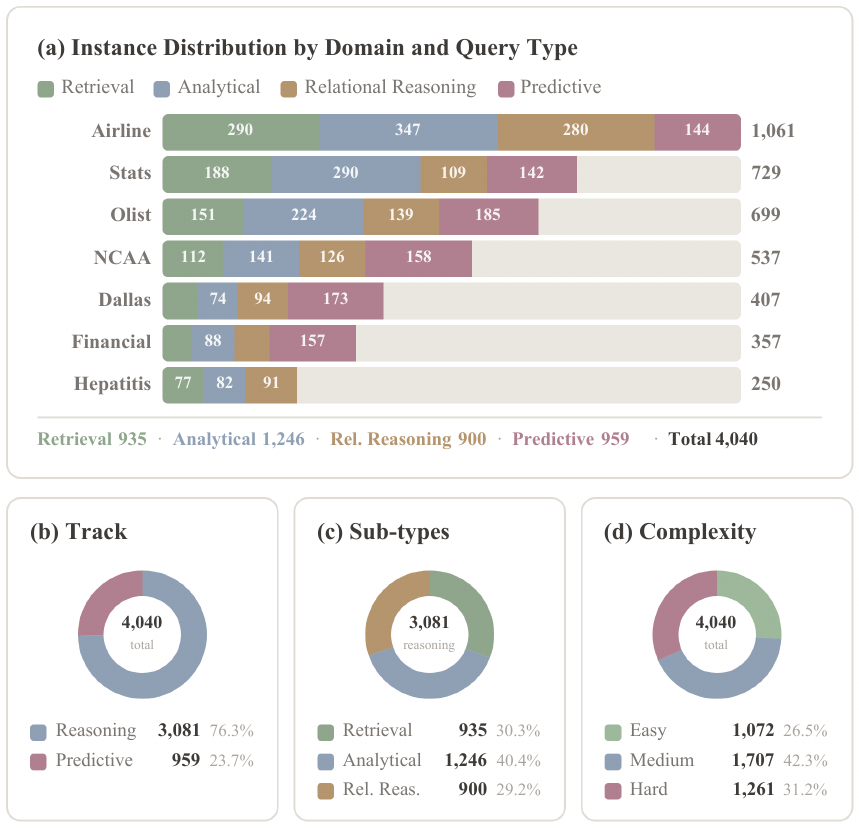}
  \vspace{-2.5em}
  \caption{Dataset statistics of RDL-Bench.
  (a) Instance distribution by domain and query type. (b) Track distribution. (c) Reasoning sub-type breakdown. (d) Complexity distribution.
  }
  \vspace{-2.2em}
  \label{fg:dataset_stats}
\end{figure}

\begin{table*}[t]
\centering
\scriptsize
\caption{Baseline methods in RDL-Bench.}
\vspace{-1em}
\label{tab:baseline_applicability}
\resizebox{\textwidth}{!}{%
\begin{tabular}{llccll}
\toprule
\textbf{Family} & \textbf{Method} & 
\textbf{Reasoning} & \textbf{Predictive} & 
\textbf{Input} & \textbf{Role} \\
\midrule
\multirow{4}{*}{LLM-based}
& LLM-Schema   
& \checkmark & --         
& Question, schema
& Schema-only reasoning solver \\
& LLM-2Hop     
& \checkmark & --         
& Question, schema, 2-hop subgraph
& Context-augmented reasoning solver \\
& LLM-only     
& --         & \checkmark 
& Schema, profiling statistics
& Data-blind predictive reference \\
& LLM-GraphCtx     
& --         & \checkmark 
& Schema, time-filtered 2-hop subgraph
& Graph-context predictive reference \\
\midrule
\multirow{5}{*}{Text-to-SQL}
& DIN-SQL~\cite{pourreza2023din}      
& \checkmark & --         
& Question, schema
& Prompting-based SQL parser \\
& C3~\cite{dong2023c3}           
& \checkmark & --         
& Question, schema
& Zero-shot SQL parser \\
& RESDSQL~\cite{li2023resdsql}      
& \checkmark & --         
& Question, schema
& Fine-tuned SQL parser \\
& CHESS~\cite{talaei2024chess}
& \checkmark & --
& Question, schema, database
& Multi-agent SQL synthesis \\
& XiYanSQL-32B~\cite{liu2026xiyan}
& \checkmark & --
& Question, schema
& Specialized SQL generation model \\
\midrule
\multirow{2}{*}{Tabular}
& XGBoost~\cite{chen2016xgboost}      
& --         & \checkmark 
& Aggregated entity features
& Tabular predictive solver \\
& LightGBM~\cite{ke2017lightgbm}     
& --         & \checkmark 
& Aggregated entity features
& Tabular predictive solver \\
\midrule
\multirow{4}{*}{Graph / RDL}
& GraphSAGE~\cite{hamilton2017inductive}    
& --         & \checkmark 
& Relational graph
& Inductive graph baseline \\
& RelGNN~\cite{chen2025relgnn}       
& --         & \checkmark 
& Relational graph
& Relational GNN \\
& RelGT~\cite{dwivedi2025relational}        
& --         & \checkmark 
& Relational graph
& Relational graph transformer \\
& StructOpt-RDL~\cite{cheng2026makes} 
& --         & \checkmark 
& Relational graph
& Structure-optimized solver \\
\midrule
\multirow{2}{*}{Task-Aware}
& LLM-Router + Modules    
& \checkmark & \checkmark 
& Question, schema
& Task-aware modular system \\
& Oracle-Router + Modules 
& \checkmark & \checkmark 
& Ground-truth metadata
& Routing upper bound \\
\bottomrule
\end{tabular}
}
\vspace{-2em}
\end{table*}

\vspace{-0.8em}
\subsection{Instance Generation}
\label{subsec:instance_generation}

We construct benchmark instances through a structured model-assisted pipeline using Claude Sonnet 4.6 as the generation model.
The pipeline uses task-specific prompt templates for reasoning-oriented and prediction-oriented questions, as shown in Figures~\ref{fig:prompt-reasoning} and~\ref{fig:prompt-predictive}.
The reasoning prompt generates question-SQL pairs with optional temporal constraints.
The prediction prompt generates future-oriented questions and enforces a causal boundary at $t_{\mathrm{ref}}$: input context is restricted to records before $t_{\mathrm{ref}}$, while ground-truth labels are derived only from records after $t_{\mathrm{ref}}$.
The generated instances are then validated and normalized into the shared instance schema.
Validation combines automatic checks with instance-by-instance manual review, as detailed in Section~\ref{subsec:quality_control}.

\paragraph{Schema and data profiling}
For each database, we build a structured schema and profiling file.
The schema records table definitions, attributes, data types, primary and foreign key relations.
The data profile records row counts, value distributions, cardinalities, null ratios, representative values, and temporal coverage.
These profiles constrain instance generation and ensure that generated questions refer to valid tables, columns, relationships, and realistic values.

\paragraph{Candidate generation}
Given the schema and data profiles, we generate candidate instances in structured JSON format.
Instances are generated separately for retrieval, analytical, relational reasoning, and predictive questions to improve coverage across task types.
For reasoning-oriented questions, each candidate includes an NL question, task metadata, SQL evidence, and ground-truth fields.
For prediction-oriented questions, each candidate includes an NL question, task metadata, a reference time, an observable context definition, target-construction logic, and ground-truth fields.

\paragraph{Temporal handling}
For prediction-oriented questions, $t_{\mathrm{ref}}$ is mandatory and defines the context--target boundary described above.
For reasoning-oriented questions, temporal conditions are included only when required by the question.
All selected time ranges must fall within the temporal coverage of the corresponding database.

\vspace{-0.5em}
\subsection{Ground-Truth Construction}
\label{subsec:ground_truth_construction}

For reasoning-oriented questions, the model-generated SQL is automatically checked and manually inspected for NL--SQL alignment, with corrections made when necessary.
We then execute the validated query against the corresponding database to obtain the denotation-level answer.
The result is stored as \texttt{raw\_value} and normalized according to the answer type.
For large set-valued outputs, we store the complete denotation for evaluation and may include truncated examples for display.

For prediction-oriented questions, labels are constructed by applying the validated target-construction logic to future records relative to $t_{\mathrm{ref}}$.
The input features or relational context are derived only from records before $t_{\mathrm{ref}}$.
This separation prevents temporal leakage.
After constructing the structured answer, we generate a canonical formatted NL answer and optional stylistic variants.
These answers provide human-readable references.

\vspace{-0.5em}
\subsection{Quality Control}
\label{subsec:quality_control}

All candidate instances undergo automatic and manual verification to check semantic alignment and temporal validity.
Automatic validation checks schema validity, SQL executability, answer consistency, temporal constraints, metadata consistency, and output formatting.
For reasoning-oriented questions, the stored \texttt{raw\_value} must match the SQL execution result.
For prediction-oriented questions, the target-construction logic must respect $t_{\mathrm{ref}}$ and exclude future information from the observable context.
One database researcher manually inspects every candidate instance for semantic correctness and ambiguity.
We check whether the NL question matches the intended relational operation, whether the SQL query or target-construction logic implements the question correctly, and whether the formatted answer faithfully describes the raw value.
Instances with ambiguous wording, invalid joins, unrealistic values, temporal leakage, trivial outputs, or inconsistent answers are revised or removed.
Manual verification resulted in the revision or removal of 382 instances.
The final benchmark contains 4,040 validated instances.
Summary statistics of the final benchmark are reported in Figure~\ref{fg:dataset_stats}.
\footnote{The hepatitis domain contains no prediction-oriented instances because it is the only non-temporal database in RDL-Bench and has no timestamp attributes.}

\begin{figure}[h]
\centering
\footnotesize
\begin{tcolorbox}[
    colback=gray!3,
    colframe=gray!55,
    width=\columnwidth,
    boxrule=0.5pt,
    arc=1mm,
    left=1mm,
    right=1mm,
    top=1mm,
    bottom=1mm
]
\textbf{Prompt for LLM-Schema}

\textit{Input:} question, database schema.

\begin{quote}
You are given a relational database schema and a natural-language question.

Generate an executable SQL query that answers the question using only the provided schema.
Then return the final answer.

Database schema: \{schema\}

Question: \{question\}

Output format:

SQL: \{sql\}

Answer: \{answer\}
\end{quote}
\end{tcolorbox}
\vspace{-1.5em}
\caption{
Prompt template for LLM-Schema.
This baseline evaluates schema-only reasoning over the observable database state.
}
\vspace{-1.5em}
\label{fig:prompt_llm_schema}
\end{figure}

\begin{figure}[h]
\centering
\footnotesize
\begin{tcolorbox}[
    colback=gray!3,
    colframe=gray!55,
    width=\columnwidth,
    boxrule=0.5pt,
    arc=1mm,
    left=1mm,
    right=1mm,
    top=1mm,
    bottom=1mm
]
\textbf{Prompt for LLM-2Hop}

\textit{Input:} question, schema, retrieved two-hop relational context.

\begin{quote}
You are given a natural-language question, the database schema, and a two-hop relational context retrieved from tables relevant to the question.

Use the schema and the local relational context to generate an executable SQL query and answer the question.

Do not use tables or columns that are not provided.

Database schema: \{schema\}

Two-hop relational context: \{two\_hop\_context\}

Question: \{question\}

Output format:

SQL: \{sql\}

Answer: \{answer\}
\end{quote}
\end{tcolorbox}
\vspace{-1.5em}
\caption{
Prompt template for LLM-2Hop.
This baseline augments schema-only reasoning with local relational context.
}
\vspace{-2em}
\label{fig:prompt_llm_2hop}
\end{figure}

\section{Experiments and Analysis}
\label{sec:experiments}

We use RDL-Bench to analyze the capability boundaries of current relational QA methods. The goal is not only to compare solvers, but also to understand which parts of task-aware relational QA remain unsolved. We therefore evaluate methods at two levels. The native-solver setting evaluates each method on the capability tracks matching its intended interface, isolating execution performance. The modular-QA setting evaluates whether a router can infer the required capability from the natural-language question and dispatch it to the appropriate execution module, exposing system-level routing errors.

The analysis is organized around four research questions. \textbf{RQ1} asks why current methods still fail on denotation-level relational reasoning. \textbf{RQ2} asks whether LLMs actually perform relational prediction or mainly exploit task-format priors. \textbf{RQ3} asks what information task-aware routing must recover beyond schema grounding. \textbf{RQ4} asks how difficulty, answer type, and domain properties explain the observed failure modes.

\begin{table*}[t]
\centering
\caption{
Native Reasoning Track results by domain.
Each domain score averages per-instance scores using
normalized EM for Scalar, Set-F1 for Set, Accuracy for
Boolean, and NDCG@5 for Ranking answers.
All scores are in $[0,1]$, higher is better.
The final column reports the unweighted mean over the
seven domains.
}
\vspace{-0.8em}
\label{tab:native_reasoning_domain}
\resizebox{\textwidth}{!}{%
\begin{tabular}{llcccccccc}
\toprule
Family & Method & Olist & Financial & NCAA & Hepatitis
       & Airline & Stats & Dallas & Avg. \\
\midrule
\multirow{6}{*}{\textit{LLM-based}}
& Claude-Schema
& \textbf{0.65} & 0.45 & 0.49 & 0.50
& 0.47 & 0.42 & \textbf{0.44} & 0.49 \\
& Claude-2Hop
& 0.64 & 0.47 & 0.44 & 0.41
& 0.48 & 0.43 & 0.43 & 0.47 \\
& Qwen32B-Schema
& 0.57 & \textbf{0.48} & 0.51 & 0.41
& 0.46 & 0.42 & 0.43 & 0.47 \\
& Qwen32B-2Hop
& 0.58 & 0.40 & 0.46 & 0.41
& 0.36 & 0.45 & 0.31 & 0.42 \\
& Llama8B-Schema
& 0.38 & 0.32 & 0.31 & 0.37
& 0.35 & 0.34 & 0.21 & 0.33 \\
& Llama8B-2Hop
& 0.37 & 0.35 & 0.48 & 0.41
& 0.36 & 0.34 & 0.21 & 0.36 \\
\midrule
\multirow{5}{*}{\textit{Text-to-SQL}}
& DIN-SQL
& 0.48 & 0.42 & 0.66 & 0.23
& 0.49 & 0.46 & 0.16 & 0.41 \\
& C3
& 0.58 & 0.33 & 0.50 & 0.31
& 0.41 & 0.42 & 0.18 & 0.39 \\
& RESDSQL
& 0.25$\pm$0.04 & 0.31$\pm$0.02 & 0.52$\pm$0.03
& 0.28$\pm$0.01 & 0.36$\pm$0.03 & 0.30$\pm$0.02
& 0.15$\pm$0.01 & 0.31$\pm$0.01 \\
& XiYanSQL-32B
& 0.55 & 0.47 & 0.70 & 0.51
& 0.48 & 0.48 & 0.36 & 0.51 \\
& CHESS
& 0.56 & 0.47 & \textbf{0.72} & \textbf{0.55}
& \textbf{0.53} & \textbf{0.50} & 0.41 & \textbf{0.53} \\
\bottomrule
\end{tabular}
}
\vspace{-2em}
\end{table*}

\subsection{Experimental Setup}

\paragraph{Baselines}
Table~\ref{tab:baseline_applicability} summarizes 
baseline applicability and key characteristics.
We evaluate four categories of models.

\begin{itemize}
    \item \textbf{LLM-based Agents.}
    Claude Sonnet 4.6~\cite{bai2022constitutional},
    Qwen2.5-32B-Instruct~\cite{qwen}, and 
    Llama-3.1-8B-Instruct~\cite{grattafiori2024llama} 
    represent closed-source large, open-source large, 
    and open-source small LLMs respectively, spanning 
    a wide capability range.
    We define the LLM-based reference configurations
    for this benchmark to isolate schema-only reasoning,
    local relational context, aggregate historical signal,
    and task-aware routing under a shared NL interface.
    On the Reasoning Track, \textbf{LLM-Schema} prompts 
    each model to generate SQL directly from the 
    question and full database schema, while 
    \textbf{LLM-2Hop} augments the schema with a 
    retrieved two-hop relational context to reduce 
    schema grounding difficulty.
    On the Predictive Track, \textbf{LLM-only} provides 
    each model with only the schema and aggregate 
    profiling statistics over $D_{\mathrm{ctx}}$, 
    serving as a data-blind reference that tests whether 
    statistical priors alone provide predictive signal.
    \textbf{LLM-GraphCtx} extends LLM-only by providing 
    a time-filtered two-hop relational subgraph constructed 
    from $D_{\mathrm{ctx}}$, i.e., only records with 
    $t(\tau) \leq t_{\mathrm{ref}}$ are included.
    This baseline tests whether access to raw relational 
    records, rather than aggregate statistics alone, provides 
    additional predictive signal for LLMs, while strictly 
    respecting the temporal boundary.

    \item \textbf{Text-to-SQL Models.}
    \textbf{DIN-SQL}~\cite{pourreza2023din} decomposes 
    complex SQL generation into sub-problems via 
    chain-of-thought prompting and self-correction, 
    achieving strong performance on multi-table queries.
    \textbf{C3}~\cite{dong2023c3} improves zero-shot 
    SQL generation through calibration and consistency 
    checking, without task-specific fine-tuning.
    \textbf{RESDSQL}~\cite{li2023resdsql} decouples 
    schema linking from SQL skeleton parsing via a 
    two-stage pipeline, and is fine-tuned on the 
    RDL-Bench training split.
    We additionally evaluate \textbf{CHESS}~\cite{talaei2024chess},
    a multi-agent SQL synthesis framework, and
    \textbf{XiYanSQL-QwenCoder-32B}~\cite{liu2026xiyan},
    a model specialized for SQL generation,
    on Olist, NCAA, Financial, and Hepatitis.

    \item \textbf{Tabular and Graph-based RDL Models.}
    \textbf{XGBoost}~\cite{chen2016xgboost} and 
    \textbf{LightGBM}~\cite{ke2017lightgbm} operate on 
    flattened entity-level feature vectors aggregated 
    from $D_{\mathrm{ctx}}$, representing the standard 
    tabular learning approach for relational prediction.
    \textbf{GraphSAGE}~\cite{hamilton2017inductive} 
    models relational structure directly via inductive 
    neighborhood aggregation, serving as a general 
    graph baseline.
    \textbf{RelGNN}~\cite{chen2025relgnn} extends 
    message passing with composite relational operators 
    designed specifically for multi-table RDB structure.
    \textbf{RelGT}~\cite{dwivedi2025relational} applies 
    a graph transformer architecture to heterogeneous 
    relational graphs, capturing long-range dependencies 
    across tables.
    \textbf{StructOpt-RDL}~\cite{cheng2026makes} 
    optimizes the relational graph structure through 
    information filtering and structural injection 
    before message passing.

    \item \textbf{Task-Aware Systems.}
    \textbf{LLM-Router} instantiates each LLM as a 
    router that predicts 
    $z{=}\{track,\ query\_type,\ answer\_nature,\ 
    tables\_involved\}$ from the question and schema 
    alone, without access to ground-truth metadata.
    \textbf{Oracle-Router} uses ground-truth instance 
    metadata to always select the correct track and 
    execution module, providing a routing upper bound 
    that isolates execution performance from routing 
    errors.
\end{itemize}

\begin{table*}[t]
\centering
\caption{
Reasoning Track performance by answer type.
Primary and secondary metrics are reported for each 
answer type.
All metrics except nMAE are in $[0,1]$, higher is 
better.
AUROC is not applicable (--) for LLM-based methods 
as they output discrete Yes/No without confidence 
scores.
}
\vspace{-1em}
\label{tab:reasoning_answer_type}
\resizebox{\textwidth}{!}{%
\begin{tabular}{ll
    cc
    cc
    cc
    cc}
\toprule
& & 
\multicolumn{2}{c}{Scalar} &
\multicolumn{2}{c}{Set} &
\multicolumn{2}{c}{Boolean} &
\multicolumn{2}{c}{Ranking} \\
\cmidrule(lr){3-4}\cmidrule(lr){5-6}
\cmidrule(lr){7-8}\cmidrule(lr){9-10}
Family & Method & 
EM$\uparrow$ & nMAE$\downarrow$ & 
F1$\uparrow$ & EM$\uparrow$ & 
Acc$\uparrow$ & AUROC$\uparrow$ & 
NDCG@5$\uparrow$ & $\rho\uparrow$ \\
\midrule
\multirow{6}{*}{LLM-based}
& Claude-Schema    
& 0.76 & 0.51 
& 0.62 & 0.41 
& \textbf{0.83} & -- 
& 0.01 & 0.01 \\
& Claude-2Hop      
& \textbf{0.81} & 0.43
& 0.52 & 0.35 
& 0.80 & -- 
& \textbf{0.05} & \textbf{0.04} \\
& Qwen32B-Schema   
& 0.76 & 0.10 
& 0.32 & 0.19 
& 0.66 & -- 
& 0.03 & 0.02 \\
& Qwen32B-2Hop     
& 0.76 & 0.94 
& 0.32 & 0.20 
& 0.65 & -- 
& \textbf{0.05} & \textbf{0.04} \\
& Llama8B-Schema   
& 0.61 & 1.18 
& 0.26 & 0.14 
& 0.50 & -- 
& 0.03 & 0.02 \\
& Llama8B-2Hop     
& 0.61 & 6.50 
& 0.26 & 0.15 
& 0.50 & -- 
& \textbf{0.05} & \textbf{0.04} \\
\midrule
\multirow{3}{*}{Text-to-SQL}
& DIN-SQL          
& 0.79 & \textbf{0.07} 
& 0.31 & 0.18 
& 0.53 & 0.61 
& 0.01 & \textbf{0.01} \\
& C3               
& 0.75 & 0.01 
& \textbf{0.83} & \textbf{0.65} 
& 0.55 & 0.54 
& 0.03 & 0.02 \\
& RESDSQL          
& $0.74\pm0.02$ & $0.28\pm0.03$ 
& $0.70\pm0.02$ & $0.52\pm0.02$ 
& $0.41\pm0.02$ & $0.49\pm0.03$ 
& $0.01\pm0.01$ & $0.01\pm0.01$ \\
\bottomrule
\end{tabular}
}
\vspace{-2em}
\end{table*}

\paragraph{Implementation Details}
\label{subsec:implementation_details}

\begin{itemize}

\item \textbf{LLM Configuration.}
All LLM-based components use 3-shot in-context
learning with examples drawn from the training
partition. 
LLM-2Hop retrieves candidate tables via
BM25 matching between question tokens and column
descriptions, expands along foreign-key edges up to
two hops, and includes up to three representative rows
per table as compact record samples. 
LLM-only receives
the schema and a data profiling summary computed
exclusively over $D_{ctx}$, covering row counts, value
distributions, and temporal coverage up to
$t_{\text{ref}}$; no statistics beyond $t_{\text{ref}}$
are exposed. 
LLM-Router requires structured JSON output
$z$; responses that fail to parse
default to the Reasoning Track.
Figures~\ref{fig:prompt_llm_schema} and Figure~\ref{fig:prompt_llm_2hop}
show the prompt templates used for LLM-based baselines.
Figure~\ref{fig:prompt_llm_router} shows the prompt templates used for LLM-Router.
For each LLM-based component, the same prompt template is used across Claude, Qwen32B, and Llama8B under deterministic decoding.

\item \textbf{Reasoning Solver Implementation.}
All generated SQL is executed via DuckDB; non-executable
queries are marked incorrect. DIN-SQL and C3 use
few-shot prompting. RESDSQL is fine-tuned on the
training partition with early stopping on the
validation partition.

\item \textbf{Predictive Solver Implementation.}
For trained predictive baselines, each Predictive Track instance is instantiated as an entity-level supervised task using the oracle task specification from the benchmark metadata.
For each predictive template, we sample up to \(n=5{,}000\) anchor entities and split them into 60\% training, 20\% validation, and 20\% test examples within the template.
All features are computed exclusively from \(D_{\mathrm{ctx}}\), and no records after \(t_{\mathrm{ref}}\) are used for feature construction.
Tabular models construct entity-level feature vectors using aggregate statistics over related tables, including counts, sums, means, extrema, categorical frequencies, distinct counts, and recency-based features.
XGBoost and LightGBM are trained separately for each instantiated predictive template with validation-based model selection.
Graph-based models convert \(D_{\mathrm{ctx}}\) into a heterogeneous relational graph, where each tuple is represented as a node and foreign-key relations are represented as typed edges.
Anchor entities are mapped to their corresponding tuple nodes and serve as supervised prediction nodes.
Labels are generated from future records according to the target-construction logic.
GraphSAGE, RelGNN, and RelGT perform \(k=2\) hop message passing and attach a task-specific classification or regression head to the anchor-node representation.
All neural predictive models are trained with early stopping on the validation split.

\item \textbf{Task-Aware System Implementation.}
The execution modules are fixed as the
best-performing native solvers from each track on the
validation split. Oracle-Router uses ground-truth
\texttt{query\_type} to determine the capability
track.

\item \textbf{Environment.}
All SQL execution and data preparation use DuckDB.
Neural models are implemented in PyTorch and trained
on NVIDIA A30 GPUs. All results are averaged over
five independent runs with different seeds.

\end{itemize}


\begin{table*}[t]
\centering
\caption{
Native Predictive Track classification results by domain.
Each cell reports Accuracy over Boolean Predictive 
tasks instances.
The final column reports the unweighted macro average 
over the six temporal domains.
Higher is better.
}
\vspace{-1em}
\label{tab:native_predictive_cls}
\resizebox{\textwidth}{!}{%
\begin{tabular}{llccccccc}
\toprule
Family & Method & Olist & Financial & NCAA 
       & Airline & Stats & Dallas & Macro Avg. \\
\midrule
\multirow{7}{*}{LLM-based}
& Claude-only 
& 0.88 & 0.72 & 0.61 & 0.52 
& \textbf{0.91} & 0.63 & 0.71 \\
& Qwen32B-only 
& 0.66 & 0.64 & 0.51 & 0.54 
& 0.59 & 0.57 & 0.59 \\
& Llama8B-only 
& 0.74 & 0.53 & 0.50 & 0.44 
& 0.60 & 0.58 & 0.57 \\
\cmidrule{2-9}
& Claude-GraphCtx
& \textbf{0.89} & 0.74 & 0.65 & 0.55
& 0.92 & 0.67 & 0.74 \\
& Qwen32B-GraphCtx
& 0.68 & 0.66 & 0.55 & 0.56
& 0.61 & 0.61 & 0.61 \\
& Llama8B-GraphCtx
& 0.75 & 0.55 & 0.53 & 0.46
& 0.61 & 0.62 & 0.59 \\
\midrule
\multirow{2}{*}{Tabular}
& XGBoost 
& $0.78\pm0.02$ & $0.53\pm0.03$ & $0.54\pm0.01$ 
& $0.53\pm0.02$ & $0.55\pm0.02$ & $0.70\pm0.01$ 
& 0.60 \\
& LightGBM 
& $0.78\pm0.02$ & $0.58\pm0.03$ & $0.58\pm0.01$ 
& $0.41\pm0.02$ & $0.66\pm0.03$ & $0.80\pm0.02$ 
& 0.63 \\
\midrule
\multirow{4}{*}{RDL}
& GraphSAGE 
& $0.55\pm0.02$ & $0.86\pm0.02$ 
& $\mathbf{0.75\pm0.02}$ & $0.71\pm0.02$ 
& $0.57\pm0.04$ & $0.86\pm0.03$ & 0.72 \\
& RelGNN 
& $0.85\pm0.02$ & $\mathbf{0.93\pm0.01}$ 
& $0.52\pm0.01$ & $0.64\pm0.01$ 
& $0.62\pm0.01$ & $\mathbf{0.94\pm0.02}$ & 0.75 \\
& RelGT 
& $0.65\pm0.02$ & $0.92\pm0.01$ 
& $0.62\pm0.01$ & $0.64\pm0.01$ 
& $0.65\pm0.01$ & $0.93\pm0.02$ & 0.73 \\
& StructOpt-RDL 
& $0.86\pm0.01$ & $0.91\pm0.01$ 
& $0.73\pm0.01$ & $\mathbf{0.72\pm0.01}$ 
& $0.73\pm0.01$ & $0.64\pm0.02$ 
& $\mathbf{0.77}$ \\
\bottomrule
\end{tabular}
}
\vspace{-1.8em}
\end{table*}


\paragraph{Metrics}
We evaluate answers against the denotation-level ground truth stored in \texttt{raw\_value} and routing decisions against ground-truth task metadata.
Results are reported separately for reasoning, prediction, and routing.

\begin{itemize}
\item \textbf{Reasoning metrics.}
For reasoning-oriented questions, metrics are selected
according to the answer
type~\cite{rajpurkar2016squad,pasupat2015compositional,
jarvelin2002cumulated}.
For \textit{scalar} answers, we use normalized exact
match (EM) with a tolerance margin, and additionally
report nMAE to quantify prediction error magnitude.
For \textit{set-valued} answers, we report Set-F1 as the primary metric and Set-EM as a stricter exact-match criterion.
For \textit{boolean} answers, we report Accuracy.
For \textit{ranking} answers, we report NDCG@5.
Denotation-level answer correctness is the primary criterion throughout.

In Table~\ref{tab:native_reasoning_domain}, each instance is scored using the primary metric for its answer type, with scores in $[0,1]$.
We average these scores within each domain and report the unweighted mean over the seven domains in the final column.
This aggregation summarizes workload performance, while the underlying metrics measure different aspects of answer quality.
Table~\ref{tab:reasoning_answer_type} reports results separately by answer type.

\item \textbf{Predictive metrics.}
For the Predictive Track, classification and regression 
are evaluated 
separately~\cite{fawcett2006introduction,hyndman2006another}.
For binary classification tasks, we report Accuracy for all methods, because LLM-based predictors output discrete Yes/No answers without calibrated confidence scores. 
This keeps LLM-based and RDL predictors comparable under the same answer-level criterion.
For scalar regression tasks, all methods are evaluated 
with nMAE.

\item \textbf{Routing metrics.}
For task-aware routing, we report Track Accuracy, 
Query-Type Macro-F1, Answer-Nature Accuracy, Table-F1, 
and Module 
Accuracy~\cite{sokolova2009systematic}.
Track Accuracy measures whether the router correctly 
predicts the required capability track.
Query-Type Macro-F1 evaluates fine-grained query-type 
prediction.
Answer-Nature Accuracy measures whether the router 
predicts the expected answer form.
Table-F1 evaluates schema grounding by comparing 
predicted and ground-truth involved tables.
Module Accuracy measures whether the predicted route 
selects the correct execution module.
\end{itemize}

\begin{table*}[t]
\centering
\caption{
Native Predictive Track regression results by domain.
Each cell reports nMAE over Scalar Predictive Track 
instances instantiated as entity-level regression tasks.
The final column reports the unweighted macro average 
over the six temporal domains.
Lower is better.
}
\vspace{-1em}
\label{tab:native_predictive_reg}
\resizebox{\textwidth}{!}{%
\begin{tabular}{llccccccc}
\toprule
Family & Method & Olist & Financial & NCAA 
       & Airline & Stats & Dallas & Macro Avg. \\
\midrule
\multirow{7}{*}{LLM-based}
& Claude-only 
& 0.99 & 0.97 & 11.07 & $\mathbf{0.66}$ 
& 1.43 & 1.11 & 2.70 \\
& Qwen32B-only 
& 1.08 & 3.35 & $\mathbf{0.47}$ & 1.05 
& 1.61 & 1.15 & 1.45 \\
& Llama8B-only 
& 0.94 & 1.00 & 0.56 & 50.04 
& 1.88 & 0.94 & 9.23 \\
\cmidrule{2-9}
& Claude-GraphCtx
& 0.97 & 0.94 & 10.21 & 0.69
& 1.38 & 1.05 & 2.54 \\
& Qwen32B-GraphCtx
& 1.05 & 3.10 & 0.48 & 1.02
& 1.55 & 1.10 & 1.38 \\
& Llama8B-GraphCtx
& 0.92 & 0.98 & 0.54 & 48.73
& 1.82 & 0.91 & 8.98 \\
\midrule
\multirow{2}{*}{Tabular}
& XGBoost 
& $1.09\pm0.04$ & $0.91\pm0.03$ & $0.99\pm0.03$ 
& $11.31\pm0.06$ & $0.44\pm0.03$ 
& $\mathbf{0.48\pm0.03}$ & 2.54 \\
& LightGBM 
& $1.00\pm0.06$ & $0.92\pm0.04$ & $0.99\pm0.03$ 
& $11.04\pm0.07$ & $1.44\pm0.04$ 
& $0.66\pm0.04$ & 2.67 \\
\midrule
\multirow{4}{*}{RDL}
& GraphSAGE 
& $0.80\pm0.04$ & $0.72\pm0.02$ & $0.98\pm0.02$ 
& $0.71\pm0.04$ & $0.92\pm0.04$ 
& $0.55\pm0.05$ & 0.78 \\
& RelGNN 
& $1.00\pm0.05$ & $0.89\pm0.03$ & $1.00\pm0.02$ 
& $0.66\pm0.04$ & $\mathbf{0.30\pm0.06}$ 
& $0.50\pm0.05$ & $\mathbf{0.73}$ \\
& RelGT 
& $\mathbf{0.79\pm0.04}$ & $\mathbf{0.64\pm0.02}$ 
& $0.97\pm0.04$ & $0.81\pm0.03$ 
& $0.88\pm0.06$ & $0.67\pm0.05$ & 0.79 \\
& StructOpt-RDL 
& $0.82\pm0.03$ & $0.96\pm0.02$ & $0.98\pm0.02$ 
& $0.74\pm0.03$ & $0.63\pm0.03$ 
& $0.52\pm0.03$ & 0.78 \\
\bottomrule
\end{tabular}
}
\vspace{-2.5em}
\end{table*}



\begin{table}[t]
\centering
\scriptsize
\caption{
Cross-generator Reasoning evaluation on Olist and NCAA.
Rows denote question generators and columns denote solvers.
Higher scores are better.
}
\vspace{-1em}
\label{tab:cross_generator}
\begin{tabular}{lcc}
\toprule
Generator & Claude-Schema & Qwen32B-Schema \\
\midrule
Claude & 0.57 & 0.54 \\
Qwen   & 0.58 & 0.51 \\
\bottomrule
\end{tabular}
\vspace{-3.5em}
\end{table}

\begin{figure}[t]
\centering
\footnotesize
\begin{tcolorbox}[
    colback=gray!3,
    colframe=gray!55,
    width=\columnwidth,
    boxrule=0.5pt,
    arc=1mm,
    left=1mm,
    right=1mm,
    top=1mm,
    bottom=1mm
]
\textbf{Prompt for LLM-Router}

\textit{Input:} question, database schema.

\begin{quote}
You are given a natural-language question over a relational database and the database schema.

Decide which capability is required to answer the question.

Use \texttt{Reasoning} if the answer can be derived from the observable database state.

Use \texttt{Predictive} if the question asks for a future or unobserved outcome beyond the observable database state.

Also identify the query type, answer nature, and involved tables.

Database schema: \{schema\}

Question: \{question\}

Output format:

\{
\quad "track": "Reasoning" or "Predictive",

\quad "query\_type": \{query\_type\},

\quad "answer\_nature": \{answer\_nature\},

\quad "tables": [\{tables\}]
\}
\end{quote}
\end{tcolorbox}
\vspace{-1.5em}
\caption{
Prompt template for LLM-Router.
This setting evaluates whether an LLM can recover task route metadata from a natural-language question.
}
\vspace{-2em}
\label{fig:prompt_llm_router}
\end{figure}

\vspace{-0.8em}
\subsection{RQ1: Native Reasoning over Observable Database States}

Table~\ref{tab:native_reasoning_domain} reports
native Reasoning Track performance across seven
domains.
The best macro-average is 0.49 (Claude-Schema),
confirming that denotation-level relational reasoning
remains far from saturated for all evaluated methods.
For Claude-Schema, 94.7\% of outputs yield executable SQL;
among these executable queries, only 52.1\% return the
exact gold denotation.
The error audit identifies failures in schema grounding,
joins, filters, aggregations, and relational composition.
These results indicate that semantic alignment remains
a substantial bottleneck even when SQL execution succeeds.

Solver advantages vary across domains.
DIN-SQL outperforms all LLM-based methods on NCAA,
Airline, and Stats, but drops to 0.16 on Dallas,
showing that its decomposition-based pipeline is
not uniformly effective across database workloads.
CHESS and XiYanSQL-32B improve over DIN-SQL on
all four domains evaluated for these two methods.
CHESS achieves the highest reported scores on NCAA
(0.72) and Hepatitis (0.55), while Claude-Schema
remains strongest on Olist (0.65) and Qwen32B-Schema
on Financial (0.48).
Including these newer systems therefore strengthens
the SQL baselines while preserving the domain-dependent
pattern of solver advantages.
RESDSQL obtains the lowest macro-average among
the solvers evaluated on all seven domains despite
fine-tuning on the RDL-Bench training split,
suggesting that supervised schema-linking and
SQL-skeleton parsing do not transfer robustly
to this heterogeneous relational QA setting.
Context augmentation provides no consistent benefit:
Claude-2Hop and Qwen32B-2Hop both improve on some
domains but degrade on others, with macro-averages
below their Schema-only counterparts.
Retrieved context helps value grounding but
introduces distractors when the challenge lies in
query planning or relational composition.

Table~\ref{tab:reasoning_answer_type} further shows
that aggregate domain scores mask substantial
variation across answer forms.
Scalar answers are relatively tractable
(Claude-2Hop: 0.81 EM), while Ranking answers are
near zero for almost all methods, indicating that
ordered relational outputs require a level of
comparison and sorting that no current method
reliably provides.
C3 achieves strong Set performance (F1: 0.83, EM: 0.65),
while its Boolean accuracy is close to that of DIN-SQL
(0.55 vs.\ 0.53).
Its large advantage on set-valued answers therefore
does not translate into a comparable advantage on
Boolean questions, highlighting the answer-type-specific
strengths of specialized SQL generation.

\noindent\textbf{Generator sensitivity.}
To test generator-specific bias, we generated
100 additional Reasoning questions with
Qwen2.5-32B-Instruct, with 50 each for Olist and NCAA,
using the same schema/profiling inputs, prompt,
validation protocol, and matched task distributions.
Table~\ref{tab:cross_generator} shows that
Claude-Schema remains ahead on both Claude-generated
questions (0.57 vs.\ 0.54) and Qwen-generated questions
(0.58 vs.\ 0.51).
This preliminary two-domain comparison provides
no evidence of a Claude-specific home-field advantage.

\noindent\textbf{Finding.}
The bottleneck is not missing context or SQL
expressiveness, but the joint difficulty of schema
grounding, relational composition, and answer
normalization.
Current methods trade off across answer types
rather than jointly improving.

\vspace{-0.5em}
\subsection{RQ2: Predictive QA over Future Outcomes}

Tables~\ref{tab:native_predictive_cls} 
and~\ref{tab:native_predictive_reg} reveal sharply 
different behavior across Boolean and scalar settings.
On Boolean prediction, Claude-only achieves 0.71
macro-average Accuracy using only schema and
aggregate statistics, competitive with trained
tabular models.
Claude-GraphCtx provides marginal additional gain
(0.74), indicating that raw relational records offer
limited signal beyond aggregate profiling for binary
decisions.
This competitiveness is format-driven: binary
questions reduce relational prediction to a coarse
polarity judgment that LLMs can approximate from
statistical priors without learning.

Scalar regression exposes this limitation directly.
LLM-based methods show high nMAE and severe
cross-domain instability, with Llama8B-only reaching
50.04 on Airline.
LLM-GraphCtx provides only marginal improvement
over LLM-only, showing that additional local records
alone do not close the numerical estimation gap.
Graph-based models achieve stronger aggregate
regression performance, with RelGNN obtaining the
best macro-average nMAE of 0.73.
These models use entity-level supervision to learn
task-specific predictive mappings from relational
histories, whereas the LLM references estimate outcomes
directly from prompted summaries or local records.

\noindent\textbf{Finding.}
LLMs can exploit coarse historical priors for Boolean
prediction but remain unstable on scalar regression.
Trained relational models achieve stronger aggregate
regression performance, highlighting the value of
task-specific learning over relational histories
for numerical prediction.

\begin{table}[t]
\centering
\large
\caption{
Routing diagnosis results on all test instances.
Higher is better.
}
\vspace{-0.8em}
\label{tab:routing_diagnosis}
\resizebox{\columnwidth}{!}{%
\begin{tabular}{lccccc}
\toprule
Router & Track Acc. & QType F1 & Ans. Acc. & Table F1 & Module Acc. \\
\midrule
Random-Router 
& 0.49 & 0.46 & 0.49 & 0.00 & 0.49 \\
Llama8B-Router 
& 0.29 & 0.25 & 0.29 & 0.78 & 0.29 \\
Qwen32B-Router 
& 0.95 & 0.93 & 0.95 & 0.91 & 0.95 \\
Claude-Router 
& 0.97 & 0.96 & 0.97 & 0.94 & 0.97 \\
Oracle-Router 
& 1.00 & 1.00 & 1.00 & 1.00 & 1.00 \\
\bottomrule
\end{tabular}
}
\vspace{-1em}
\end{table}

\begin{table}[t]
\centering
\normalsize
\caption{
End-to-end modular QA results with fixed execution modules.
}
\vspace{-1em}
\label{tab:e2e_modular_qa}
\resizebox{\columnwidth}{!}{%
\begin{tabular}{lcccc}
\toprule
System & Reasoning$\uparrow$  & Bool. Acc $\uparrow$  & Scalar nMAE$\downarrow$ & Track Acc.$\uparrow$  \\
\midrule
Random-Router 
& 0.20 & 0.43 & 1.06 & 0.49 \\
Llama8B-Router
& 0.03 & 0.43 & 1.91 & 0.29 \\
Qwen32B-Router
& 0.44 & 0.72 & 1.10 & 0.95 \\
Claude-Router
& 0.44 & 0.84 & 0.98 & 0.97 \\
Oracle-Router
& 0.44 & 0.87 & 0.91 & 1.00 \\
\bottomrule
\end{tabular}
}
\vspace{-2em}
\end{table}

\begin{figure*}[t]
\centering
\includegraphics[width=\textwidth]{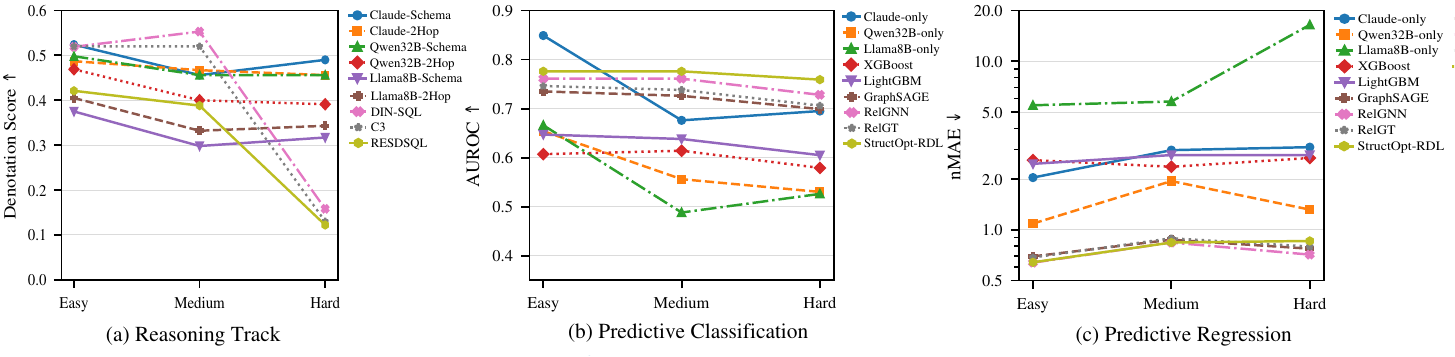}
\vspace{-2.5em}
\caption{
Performance trends across difficulty levels.
Panel (a) reports Reasoning Track denotation scores, where higher is better.
Panel (b) reports Predictive Classification AUROC, where higher is better.
Panel (c) reports Predictive Regression nMAE, where lower is better.
}
\vspace{-2em}
\label{fig:difficulty_trends}
\end{figure*}

\begin{figure}[t]
\centering
\includegraphics[width=\columnwidth]{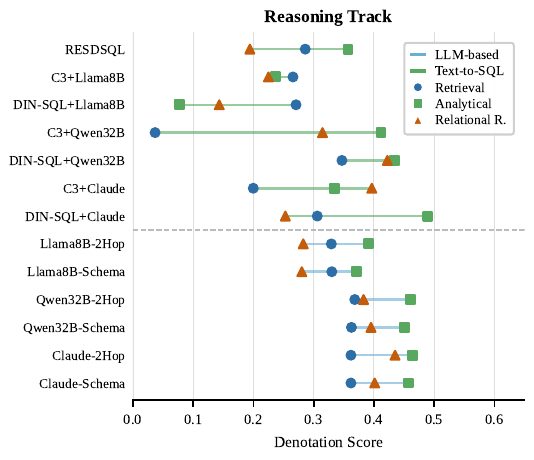}
\vspace{-2.5em}
\caption{
Reasoning Track performance by query type.
}
\vspace{-1.5em}
\label{fig:reasoning_track}
\end{figure}

\vspace{-0.5em}
\subsection{RQ3: Task-Aware Routing and End-to-End QA}

Table~\ref{tab:routing_diagnosis} evaluates routing
as a system-level diagnostic of capability understanding
within the unified NL workload.
Claude-Router and Qwen32B-Router achieve Track
Accuracy of 0.97 and 0.95, respectively, closely
approaching the Oracle upper bound.
Llama8B-Router falls to 0.29, below the
Random-Router baseline of 0.49.
A critical observation is that table grounding and
capability identification are separable skills.
Llama8B-Router achieves Table-F1 of 0.78 while
failing on Track Accuracy (0.29), showing that a
model can correctly identify relevant schema
components yet fail to determine whether the
question requires reasoning or
future-oriented prediction.
This dissociation reveals that task-aware routing
is not reducible to schema linking; it requires
understanding the temporal and predictive intent
implied by the question.

Table~\ref{tab:e2e_modular_qa} compares end-to-end
QA performance with fixed execution modules and
different routers.
Llama8B-Router drops to a Reasoning score of 0.03,
compared to 0.44 for Claude-Router and Oracle-Router.
Because the execution modules are fixed, this gap
reveals the system-level impact of incorrect
capability selection.
A track-routing error can send a future-oriented
question to a current-state query solver, or a
current-state question to a predictive model.
The resulting failure is a mismatch between the
question's required capability and the invoked
execution mechanism.
The residual gap between Claude-Router and
Oracle-Router is concentrated in the Predictive Track
(Bool. Acc.: 0.84 vs.\ 0.87; Scalar nMAE:
0.98 vs.\ 0.91).
Thus, even near-oracle Track Accuracy can leave
measurable gaps in end-to-end predictive QA.

\noindent\textbf{Finding.}
Routing diagnoses whether schema grounding is
accompanied by correct execution-capability selection.
Strong LLMs approach oracle routing, while the
Llama8B results show that identifying relevant
tables alone does not ensure effective task-aware QA.
Comparing learned and oracle routing exposes
the performance gap caused by capability-selection
errors under fixed execution modules.

\vspace{-0.5em}
\subsection{RQ4: Diagnostic Analysis}

\begin{figure}[t]
\centering
\includegraphics[width=\columnwidth]{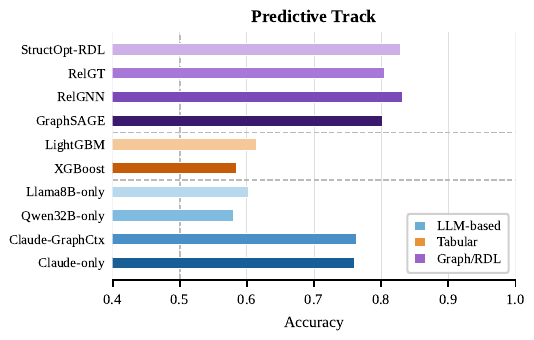}
\vspace{-2.5em}
\caption{
Predictive Track performance (Accuracy) by method.
}
\vspace{-2em}
\label{fig:predictive_track}
\end{figure}

Figures~\ref{fig:difficulty_trends}, 
\ref{fig:reasoning_track}, 
and~\ref{fig:predictive_track} provide diagnostic 
breakdowns that explain why aggregate scores alone 
are insufficient.

\paragraph{Difficulty}
Annotated difficulty is useful but not monotonic.
For Reasoning Track, Text-to-SQL methods degrade 
sharply on Hard instances while LLM-based methods 
remain relatively stable, suggesting that SQL 
decomposition fails on complex multi-hop conditions 
whereas LLMs maintain broader compositional 
coverage at the cost of lower peak performance.
For Predictive Track, difficulty trends are less 
regular, particularly for scalar regression, where 
performance depends on temporal signal strength 
and label distribution rather than structural 
complexity alone.

\paragraph{Query type}
Figure~\ref{fig:reasoning_track} shows that query-type
performance varies across methods and LLM backbones.
DIN-SQL with Claude achieves the highest Analytical
score (0.490) but drops sharply on Relational Reasoning
(0.253), revealing that decomposition-based SQL
generation handles aggregation well but struggles
with nested multi-table composition.
On the Predictive Track, graph and RDL models are
generally more reliable than LLM-based methods,
especially for scalar regression, but classification
results show that LLMs can remain competitive on
some Boolean tasks.

\noindent\textbf{Finding.}
Failure modes are heterogeneous across difficulty 
levels and query types.
Text-to-SQL methods fail on structural complexity; 
LLM-based methods fail on compositional depth; 
predictive methods fail when temporal signal is 
weak or label distributions are irregular.
No single improvement addresses all dimensions 
simultaneously.

\begin{table}[t]
\centering
\scriptsize
\caption{
Average runtime of representative solvers.
Predictive solvers are reported in seconds over the six temporal domains.
LLM-based routing and reasoning pipelines are reported in minutes over seven domains.
Lower is better.
}
\vspace{-1em}
\label{tab:runtime}
\begin{tabular}{llc}
\toprule
Family & Method & Avg. Runtime \\
\midrule
\multirow{6}{*}{Predictive}
& XGBoost & 28.3s \\
& LightGBM & \textbf{25.2s} \\
& GraphSAGE & 35.0s \\
& RelGNN & 34.4s \\
& RelGT & 34.8s \\
& StructOpt-RDL & 29.1s \\
\midrule
\multirow{3}{*}{Router}
& Claude-Router & 14.7m \\
& Qwen32B-Router & 45.0m \\
& Llama8B-Router & \textbf{12.6m} \\
\midrule
\multirow{5}{*}{Reasoning}
& Claude-Schema+2Hop & \textbf{4.0m} \\
& Qwen32B-Schema+2Hop & 15.1m \\
& Llama8B-Schema+2Hop & 5.9m \\
& Claude-DIN-SQL & 18.3m \\
& Claude-C3 & 113.7m \\
\bottomrule
\end{tabular}
\vspace{-2.5em}
\end{table}

\subsection{Efficiency Analysis}

Table~\ref{tab:runtime} reports average runtime for representative solver families.
Predictive solvers are substantially cheaper than LLM-based reasoning pipelines, with tabular methods being the fastest.
Graph and RDL methods add moderate overhead from graph construction and message passing, while remaining much cheaper than multi-call LLM-based SQL pipelines.
Among LLM methods, runtime depends strongly on model size and prompting strategy; multi-stage Text-to-SQL pipelines such as C3 are especially expensive.
These results highlight the practical importance of task-aware routing: a system should invoke expensive reasoning modules only when they are required, and use trained predictive modules for future-oriented prediction.

\vspace{-0.5em}
\section{Conclusion and Future Work}

RDL-Bench is a benchmark for task-aware question 
answering over relational databases, covering 
current-state reasoning and future-oriented 
prediction under a unified natural-language 
interface across seven real-world domains.
Our two-level evaluation characterizes native solver
performance and the system-level impact of routing errors.
Denotation-level reasoning remains far from 
saturated, with the best method reaching 0.49 
macro-average; context augmentation and SQL 
specialization each address partial failure modes 
but do not jointly solve schema grounding, 
relational composition, and answer normalization.
LLM predictive competitiveness is an artifact of 
binary task format and collapses on scalar 
regression, showing that Boolean predictive QA 
alone cannot reveal the limits of relational 
understanding.
Task-aware routing exposes a capability threshold: 
large LLMs route reliably, while small LLMs 
develop systematic bias that degrades end-to-end 
performance below random, and this degradation 
is disproportionately concentrated in predictive 
tasks where module mismatch produces fundamentally 
wrong answer types.

These findings suggest several open problems for 
the community.
Relational reasoning methods that jointly address 
schema grounding, multi-hop composition, and 
answer-form control remain absent.
Predictive QA beyond binary classification, 
including ranking, set-valued prediction, and 
scalar regression under weak temporal signal, 
is largely unsolved.
The Llama8B results motivate lightweight routers
that combine schema grounding with reliable
capability selection.
Multi-turn relational QA, where follow-up 
questions depend on previous answers or 
evolving temporal assumptions, represents a 
natural extension that RDL-Bench is designed 
to support.

\section*{Acknowledgements}
This research / project is supported by the National Research Foundation, Singapore, under its Frontier CRP Grant (NRF-F-CRP-2024-0005), and under its AI Singapore Programme (AISG Award No: AISG3-RP-2024-034). This research is also supported by Singapore MOE AcRF Tier-2 Award (MOE-T2EP20224-0005) and NTU SUG-NAP. Any opinions, findings, and conclusions or recommendations expressed in this material are those of the author(s) and do not reflect the views of the National Research Foundation, Singapore.

\section*{AI-Generated Content Acknowledgement}

During the preparation of this manuscript, the authors used Claude and GPT writing tools to improve language clarity, polish phrasing, and refine presentation. 
Claude and GPT assist with code implementation and the code has been carefully verified by the authors.
The research ideas, benchmark design, experimental setup, analysis, and conclusions were developed and verified by the authors.


{
\small
\bibliographystyle{abbrv}{\bibliography{neurips_2026}}
}



\vspace{12pt}

\color{red}
\end{document}